\documentclass{article}

\usepackage{PRIMEarxiv}

\usepackage[utf8]{inputenc} 
\usepackage[T1]{fontenc}    
\usepackage{hyperref}       
\usepackage{url}            
\usepackage{booktabs}       
\usepackage{amsfonts}       
\usepackage{nicefrac}       
\usepackage{microtype}      
\usepackage{lipsum}
\usepackage{fancyhdr}       
\usepackage{graphicx}       
\graphicspath{{media/}}     

\usepackage{amsthm} 
\usepackage{amsmath} 
\usepackage{booktabs}  
\usepackage{threeparttable}
\usepackage{multirow}
\usepackage{comment}
\usepackage{bm}
\usepackage{subcaption}
\usepackage{algpseudocode}
\usepackage[ruled,linesnumbered]{algorithm2e}

\title{ADA-GNN: Atom-Distance-Angle Graph Neural Network for Crystal Material Property Prediction
\thanks{Corresponding authors: Qianli Xing (qianlixing@jlu.edu.cn) and Bo Yang (ybo@jlu.edu.cn)}
}
\author{
  Jiao Huang \\
  Jilin University \\
  Changchun, Jilin \\
  \texttt{haungjiao20@mails.jlu.edu.cn} \\
  \And
  Qianli Xing $^{*}$\\
  Jilin University \\
  Changchun, Jilin \\
  \texttt{qianlixing@jlu.edu.cn} \\
  \And
  Jinglong Ji \\
  Jilin University \\
  Changchun, Jilin \\
  \texttt{jijl22@mails.jlu.edu.cn} \\
   \And
  Bo Yang $^{*}$\\
  Jilin University \\
  Changchun, Jilin \\
  \texttt{ybo@jlu.edu.cn} \\
}

\begin{document}
\maketitle

\begin{abstract}
Property prediction is a fundamental task in crystal material research.  
To model atoms and structures, structures represented as graphs are widely used and graph learning-based methods have achieved significant progress. 
Bond angles and bond distances are two key structural information that greatly influence crystal properties. 
However, most of the existing works only consider bond distances and overlook bond angles. 
The main challenge lies in the time cost of handling bond angles, which leads to a significant increase in inference time.
To solve this issue, we first propose a crystal structure modeling based on dual scale neighbor partitioning mechanism, which uses a larger scale cutoff for edge neighbors and a smaller scale cutoff for angle neighbors.
Then, we propose a novel \textbf{A}tom-\textbf{D}istance-\textbf{A}ngle \textbf{G}raph \textbf{N}eural \textbf{N}etwork (ADA-GNN) for property prediction tasks, which processes node information and structural information separately. 
The accuracy of predictions and inference time are improved with the dual scale modeling and the specially designed architecture of ADA-GNN.
The experimental results validate that our approach achieves state-of-the-art results in two large-scale material benchmark datasets on property prediction tasks.
\end{abstract}


\section{Introduction}

Property prediction is a fundamental task in crystal research and crucial for discovering new materials with desired characteristics \cite{choudhary2021atomistic,chen2019graph,louis2020graph}. The properties of a crystal are dictated by the types of atoms and the structural details. The atoms have intrinsic features, and structural details encompass the bond distances and bond angles between each atom \cite{gasteiger2020directional,callister2007materials,kittel2005introduction}. Thus, the key challenge of predicting crystal properties lies in effectively modeling both atomic attributes and structural information.

To address this challenge, CGCNN \cite{xie2018crystal} initially represents the crystal structure as an undirected multigraph, where nodes signify atoms within the unit cell and edges denote connections between atoms. 
Multigraph modeling of crystals has been validated in crystal property prediction tasks. 
Subsequent research has predominantly focused on refining representations based on this foundation \cite{chen2019graph,yan2022periodic}. 
Among them, PotNet \cite{lin2023efficient} achieves state-of-the-art performance by encompassing the complete set of potentials among all atoms through the infinite summation of distances. 

\begin{figure}[t]
    \centering
    \includegraphics[clip, trim=0 8.2cm 1cm 0, scale=0.24]{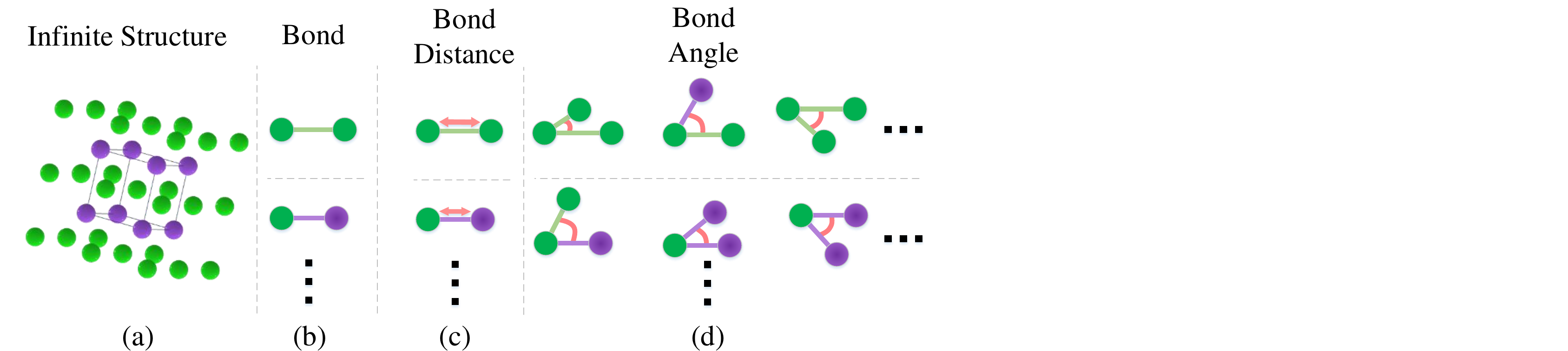}
    \caption{
    Illustrations of the crystal structure of NaCl and the bond distances and bond angles.
    (a): The structural diagram of NaCl. The green sphere signifies the atom  Cl, while the purple sphere represents the atom Na. The central cube serves as a visual representation of the unit cell.
    (b): Bonds in NaCl.
    (c): Bond distances of bonds.
    (d): Bond angles formed by the bonds.
    }
    \label{fig:intro}
\end{figure}

These prior efforts have made substantial progress by investigating in atom attributes and bond distances. However, they overlook the bond angle, which is crucial for structural information.
According to the potential energy equation between atoms \cite{gasteiger2020directional}, it is imperative to consider not only bond distances but also the structural information conveyed by bond angles, as shown below:
\begin{equation}
    E=E_{distance} + E_{angle} + E_{others}.
\end{equation}
$E_{distance}$ models on bond distances, $E_{angle}$ captures the effects of angles between bonds and $E_{others}$ encompasses interaction forces involving multiple atoms or atoms situated at a distance.
Considering that crystal properties, which are related to crystal energy, are intricately linked to these angles, it becomes imperative to incorporate angles into the modeling of the crystal structure for accuracy.

The primary difficulty in modeling bond angles arises from the infinite number of angles associated with each edge. 
For instance, as illustrated in Figure \ref{fig:intro}, considering the infinite extension of the crystal structure, each bond depicted in Figure \ref{fig:intro} (b) will have approximately an infinite number of associated angles, as shown in  Figure \ref{fig:intro} (d).
Consequently, integrating angles would lead to a significant augmentation in the number of inputs, thereby impacting the inference time. 
ALiGNN \cite{choudhary2021atomistic} divides the overall structure into point and edge graphs to solve the infinite angle problem. 
However, this model is time-consuming as
the processing of bond angles significantly increases inference time. As a result, its application in various scenarios for downstream tasks is effectively blocked. 
Thus, an effective method for processing bond angles is in need. 

To solve the infinite bond angle problems, we first propose a dual scale modeling for the crystal structure consisting of the potential energy equation.
The modeling involves utilizing a larger scale truncation for edge neighbors and a smaller scale truncation for angle neighbors, which can effectively reduce input complexity and inference time.
Then, we propose a novel \textbf{A}tom-\textbf{D}istance-\textbf{A}ngle \textbf{G}raph \textbf{N}eural \textbf{N}etwork (\textbf{ADA-GNN})  for property prediction tasks based on the modeling.
Specifically, we decouple atom attributes from structural attributes to ensure the independent embedding of atoms and structures. This approach enhances training stability and improves the accuracy of model predictions. 
The experimental results validate that our model delivers highly competitive performance in terms of inference time and achieves state-of-the-art (SOTA) results in two real-world datasets.
Our contribution mainly includes the following three parts:
\begin{itemize}

\item We propose a dual-scale modeling for crystal materials to choose edge neighbors and angle neighbors, which can leverage the information from bond angles and effectively reduce the inference time.

\item 
We propose a novel approach named ADA-GNN for property prediction tasks with a specially designed embedding process. The ADA-GNN embeds atom features and structural information separately which can help in better understanding these two kinds of information.

\item Experiments in two real-world large scale material benchmark datasets show that our approach outperforms previous state-of-the-art crystal
property prediction methods by $2.04\%-21.82\%$ in terms of MAE.
\end{itemize}

\section{Related work}
Predicting crystal properties is crucial for discovering materials with ideal properties \cite{merchant2023scaling,xie2021crystal,zhao2023physics}.
Unlike organic molecules such as small molecules \cite{wang2022advanced} and proteins \cite{morehead2021geometric,jumper2021highly}, the structure of crystal molecules is a periodic structure that extends the unit cell infinitely, which is more challenging to model \cite{ward2016general,raccuglia2016machine,oliynyk2016high}. Thus, aside from addressing the atomic types pertinent to the molecular domain, the pivotal factor in accurately predicting the quality of crystal properties lies in effectively harnessing structural information.


\begin{table}[t]  
  \centering  
  \begin{threeparttable}  
  \caption{Comparison between our method and other baselines. The columns labeled "Atom," "Distance," and "Angle" indicate whether the model incorporates atomic features, bond distance, and bond angle information, respectively. The final column, denoted as "In. Cutoff," specifies whether an independent cutoff has been established to obtain bond distances and bond angles.}  
  \label{tab:difference} 
    \begin{tabular}{c|ccc|c}
    \toprule  
    Method &Atom&Distance&Angle& In. Cutoff\cr 
    \midrule  
    CGCNN [2018]&\checkmark&\checkmark&$\times$&$\times$\cr
    SchNet [2017]&\checkmark&\checkmark&$\times$&$\times$\cr
    MEGNET [2019]&\checkmark&\checkmark&$\times$&$\times$\cr
    GATGNN [2020]&\checkmark&\checkmark&$\times$&$\times$\cr
    ALiGNN [2021]&\checkmark&\checkmark&\checkmark&$\times$\cr
    Matformer [2022]&\checkmark&\checkmark&$\times$&$\times$\cr
    PotNet [2023]&\checkmark&\checkmark&$\times$&$\times$\cr
    ADA-GNN&\checkmark&\checkmark&\checkmark&\checkmark\cr 
    \bottomrule  
    \end{tabular}  
    \end{threeparttable}  
\end{table}
Considering the unique challenges of crystal molecules, CGCNN \cite{xie2018crystal} introduces an innovative approach that incorporates both multi-graph modeling and cutoff mechanisms. 
In multigraph modeling, the neighbors of the target atom are selected based on the values of the bond distance and cutoff.
An atom neighbor is selected when the bond distance between the neighbor atom and the target atom is smaller than the cutoff. Consequently, the originally infinite-node graph structure is condensed into a finite-node graph structure.

There are a number of methods that employ the graph-structure and cutoff mechanism after the CGCNN. We compare seven representative methods, including: 
CGCNN \cite{xie2018crystal}, SchNet \cite{schutt2017schnet}, MEGNET \cite{chen2019graph}, GATGNN \cite{louis2020graph}, ALiGNN \cite{choudhary2021atomistic}, Matformer \cite{yan2022periodic}, and PotNet \cite{lin2023efficient}. All of these methods achieved the best performance during their publication period, and PotNet shows state-of-the-art performance today. 

As demonstrated in Table \ref{tab:difference}, four dimensions are employed to show their difference including atom, bond distance, bond angle, and cutoff. 
At present, our model and ALiGNN are the only two methods that incorporate angle information.
To incorporate angle details into structural information, ALiGNN uses one cutoff that partitions the entire graph structure into point and edge graphs. 
However, this approach exhibits noticeable limitations. Firstly, it overlooks the challenge of managing the quantity of bond angle information and utilizes the same scale as bond distance to gather neighbor information, resulting in a significant increase in inference time. Secondly, the continuous updating of node attributes during the modeling of structural features may impact training stability, potentially diminishing the competitiveness of the final model's predictive performance.

A significant distinction lies in our approach, where we employ two distinct cutoffs for obtaining edge and angle information, in contrast to ALiGNN, which relies solely on a single cutoff type. Additionally, unlike ALiGNN's consideration of node attributes during the modeling of structural features, our embedding module processes atom features and structure information independently.

\section{Dual Scale Modeling of Crystal Materials}

In this section, we introduce how to use graph structure to represent the crystal materials. Firstly, the existing representation of crystal materials is shown. Then, we propose a dual scale neighbor partitioning mechanism to effectively select the edge neighbors and angle neighbors. Finally, we give a problem definition of crystal property prediction tasks with the selected neighbors.

\subsection{Crystal Materials}\label{sectograph}
Crystal material is initially expressed as $\bm{M} = \{\bm A, \bm X, \bm L\}$.
$\bm A = [a_1,..., a_N]^\top \in \mathbb{A}^N$ represents atom varieties, with $\mathbb{A}$ symbolizing the set of chemical elements. 
$\bm X = [\bm x_1, ..., \bm x_N]^\top \in \mathbb{R}^{N\times3}$ specifies the spatial coordinates of atoms in the Cartesian coordinate system.
$\bm L = [\bm l_1, \bm l_2, \bm l_3]^\top \in \mathbb{R}^{3\times3}$ embodies the recurrent lattice, indicating the extending directions of unit cells in three-dimensional space. 
In this work, we represent the $M$ by using graph structures. The atoms are points and the interaction forces between atoms are edges. 
Note that the modeling of crystals must satisfy four types of invariances \cite{xie2021crystal}, including permutation invariance, translation invariance, rotation invariance, and periodic invariance.
To fulfill these requirements, absolute coordinates denoted by variables such as $X$ and $L$ must undergo a transformation into relative quantities, such as distance and angle, which are shown in the following section.

\subsection{Dual Scale Neighbor Partitioning Mechanism}\label{secdual}

\begin{figure}[t]
    \centering
    \includegraphics[clip, trim=0 1cm 0 1cm, scale=0.32]{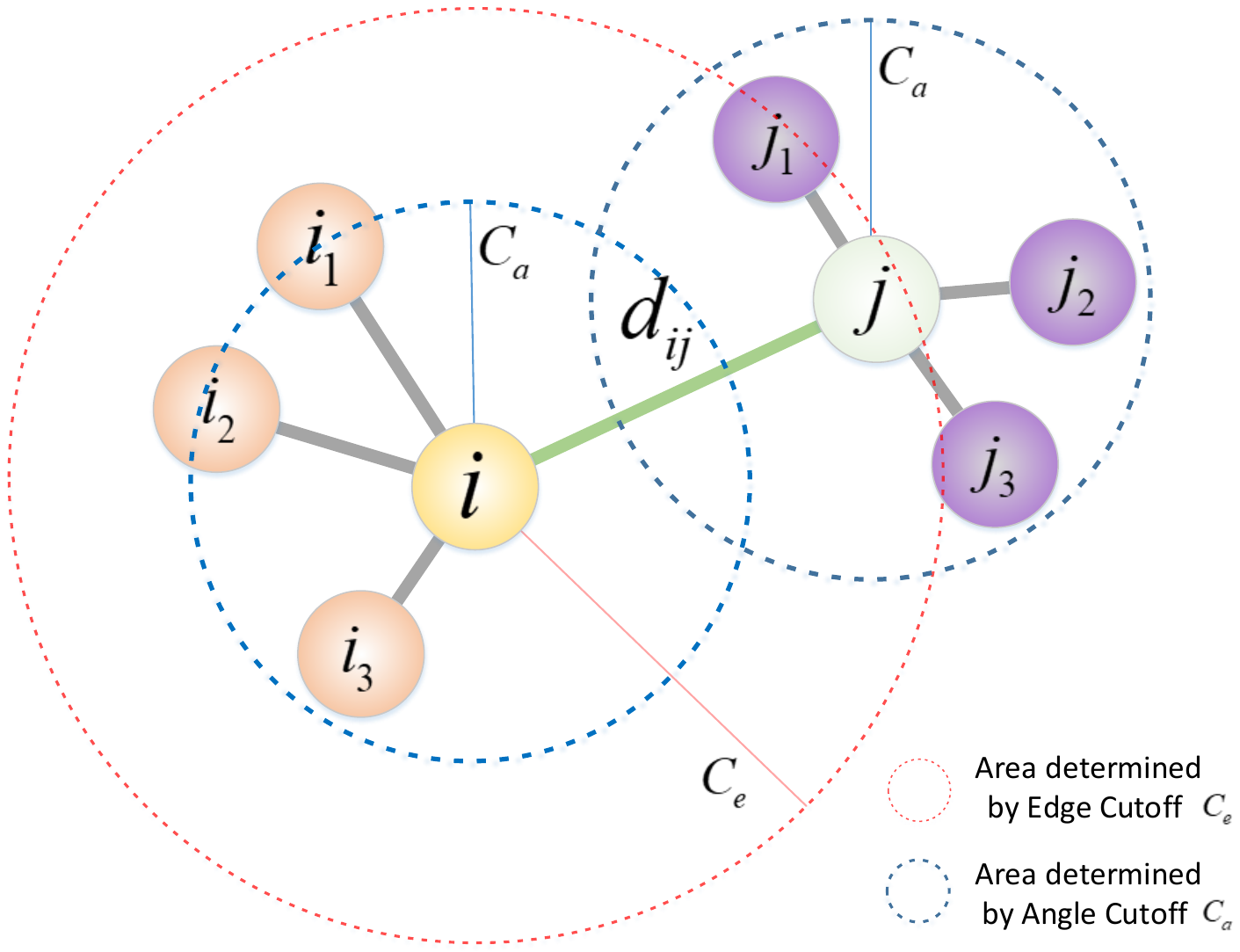}
    \caption{
    Illustrations of the partitioning mechanism for edge neighbors and angle neighbors.
    }
    \label{fig:aggregation}
\end{figure} 
In this section, we proposed a dual scale neighbor partitioning mechanism to determine the edge neighbors and angle neighbors of atoms as shown in Figure \ref{fig:aggregation}. 
Through this mechanism, we are able to represent bond distances and bond angles in a relative quantity manner.


When selecting the value of edge cutoff $C_e$, we opt for a larger scale. 
This choice is driven by two main factors: firstly, larger scales adequately fulfill the graph's connectivity prerequisites, and secondly, a sufficient number of edge neighbors at larger scales are necessary to address periodicity concerns. 
Aligning with the scale used in prior works \cite{xie2018crystal,choudhary2021atomistic} helps fulfill these criteria effectively.

For bond distances, the bond distance $d_{ij}$ between atom $i$ and atom $j$ is computed by the following formula:
\begin{equation}
    d_{i j} = ||\mathbf{x_j} - \mathbf{x_i}||_2,
\end{equation}
where atom $i$ is the atom within the unit cell and atom $j$ is atom $i$'s edge neighbor. 
If the distance $d_{ij}$ between atom $j$ and atom $i$ is less than edge cutoff $C_e$, then atom $j$ is the edge neighbor of atom $i$.
As shown in Figure \ref{fig:aggregation}, the red circle represents the area determined by edge cutoff $C_e$. For atom $i$, the atoms inside the red circle are all its edge neighbors.

When selecting the value of angle cutoff $C_a$, we opt for a smaller scale due to two primary considerations. Firstly, a smaller cutoff for angle neighbors can significantly reduce the input data volume to the network, thereby impacting inference time. We have verified this in both theoretical and experimental aspects in Section \ref{secalgorithm} and Section \ref{sectime}, respectively. Secondly, empirical findings indicate that employing smaller-scale cutoffs can still enhance algorithm performance to a state-of-the-art level.

For each edge $ij$, the angles related to the edge are divided into angles $\alpha_i$, with atom $i$ as the vertex and angles $\alpha_j$, with atom $j$ as the vertex. They are respectively computed as follows:
\begin{equation}\label{eqai}
    \alpha_i=\{\alpha_{ii_k}|\alpha_{i i_k} = angle(\overrightarrow{ii_k}, \vec{ij})\},
\end{equation}
and
\begin{equation}\label{eqaj}
    \alpha_j=\{\alpha_{jj_k}| \alpha_{jj_k} = angle(\overrightarrow{jj_k}, \vec{ji})\}.
\end{equation}
Where atom $i_k$ and atom $j_k$ are respectively the angle neighbors of atom $i$ and atom $j$. 
The angle neighbors of atoms are determined by the smaller cutoff called angle cutoff $C_a$, and the distance between atoms less than the angle cutoff $C_a$ is defined as the angle neighbors. As shown in Figure \ref{fig:aggregation}, the cyan circle represents the area determined by angle cutoff $C_a$. For atom $i$, the atoms inside the cyan circle, including atom $i_1$, atom $i_2$, and atom $i_3$, are all its angle neighbors.

Specifically, the relationship between the angle neighbor's cutoff $C_a$ and the edge neighbor's cutoff $C_e$ satisfies:
\begin{equation}\label{equneigh}
    C_e = C_a^2.
\end{equation}
Through the limitation of angle neighbor cutoff, we've successfully enhanced the model's predictive performance by leveraging angle information. Simultaneously, this restriction has effectively minimized the time impact on predictions caused by incorporating angle information.
\subsection{Crystal Material Property Prediction}
Utilizing the previously acquired edge distances $d$ and angles $\alpha$, the graph structure $G_M$ for the crystal material $M$ is expressed as follows:

\begin{equation}\label{equ_GM}
    G_M = \{A_i, d_{ij}, \alpha_{i}, \alpha_{j}\}_{\forall {i\in I}}, 
\end{equation}
where $I$ encompasses all atoms within the unit cell, and $j$ denotes the edge neighbor of atom $i$. The calculations for $\alpha_i$ and $\alpha_j$ are presented in formulas \eqref{eqai} and \eqref{eqaj}.

For a given crystal structure $M$, our goal is to predict the value of crystal energy-related properties, such as formation energy, band gap, and energy hull (ehull). Consequently, the crystal property prediction is represented as $G_M\rightarrow y\in \mathbb{R}$ in regression tasks, where $y$ denotes the crystal property value.

\section{ADA-GNN}
\begin{figure*}[t]
    \centering
    \includegraphics[clip, trim=0 9cm 0 1cm, scale=0.9]{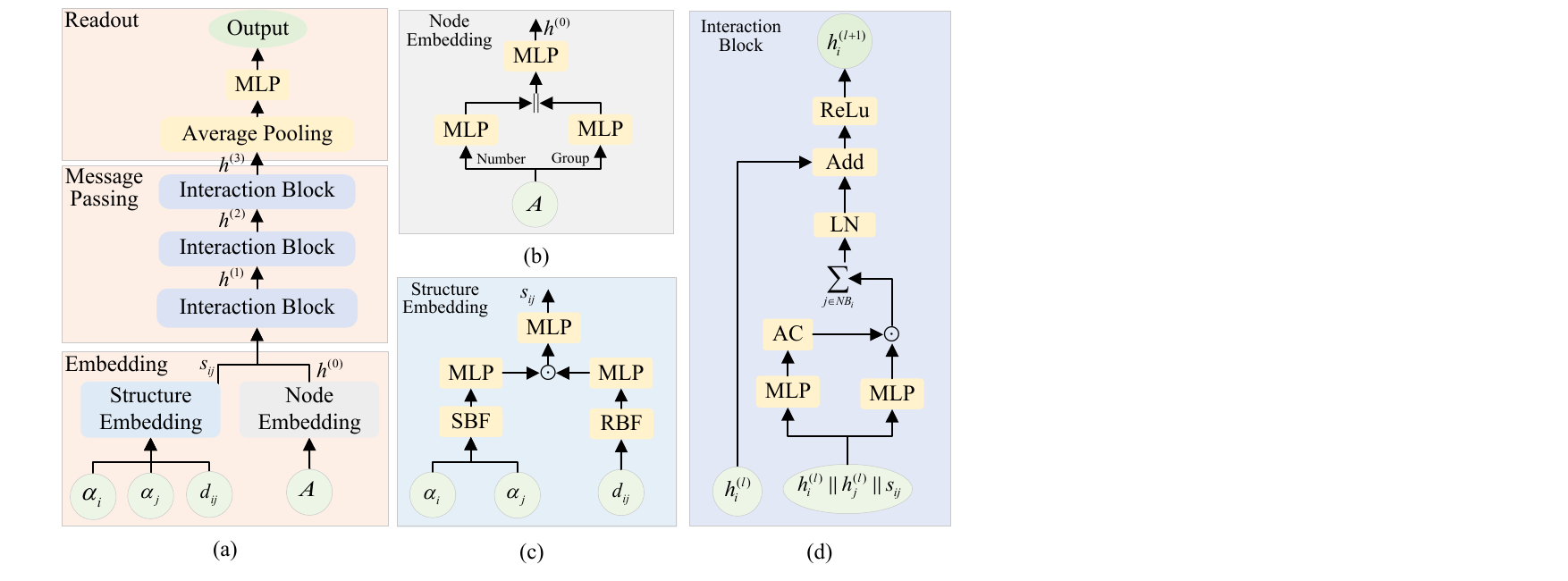}
    \caption{Illustrations of detailed Architecture of ADA-GNN. 
    (a): The overall structure of ADA-GNN.
    (b): An illustration of the node embedding module. 
    (c): An illustration of the structure embedding module. 
    (d): An illustration of the interaction block. 
    }
    \label{all_arc}
\end{figure*}

To leverage the dual scale modeling of crystal in graphs, we propose \textbf{A}tom-\textbf{D}istance-\textbf{A}ngle \textbf{G}raph \textbf{N}eural \textbf{N}etwork (\textbf{ADA-GNN}), which is composed of embedding module, message passing module and readout module
as shown in Figure \ref{all_arc}. Different from existing works, we process the atom features and structure information separately in the embedding module. Furthermore, the time cost analysis is provided to theoretically show the efficiency of the proposed approach.

\subsection{Embedding Module}
In this section, we describe how to embed the elements in $G_M$. The elements can be classified into atom attributes and structural information. Thus, we design node embedding blocks to process the atom features ($A_i$), and structure embedding blocks to process the bond distance ($d_{ij}$) and bond angles ($\alpha_{i},\alpha_{j}$). 
\subsubsection{Node Embedding Block}

As depicted in Figure \ref{all_arc}(b), the initial atom representation is computed based on their respective atom numbers and corresponding groups, which is shown as follows:

\begin{equation}\label{equnode}
\begin{aligned}
A_N &= MLP(Num_A), \\
A_G &= MLP(Grp_A),\\
h^{(0)} &= MLP(A_N \Vert A_G),
\end{aligned}
\end{equation}
Here, $Num_A$ represents the atom number, $Grp_A$ denotes the group number, and $\Vert$ signifies concatenation. Additionally, $A_N$ and $A_G$ stand for their corresponding embeddings. The assignment of group numbers is conducted manually, aligning with the configuration outlined in CGCNN \cite{xie2018crystal}. The initial atom embeddings, denoted as $h^{(0)}$, are obtained through the concatenation of $A_N$ and $A_G$.
\subsubsection{Structure Embedding Block}\label{edgeembdding}
In this section, we integrate both bond distance and angle information. This involves modeling the edge length information for each edge alongside the corresponding angle formed by that specific edge.
 

As shown in  Figure \ref{all_arc}(c), the initial structure embedding in architecture is computed as:
\begin{equation}\label{edge_embedding}
    s_{ij}=MLP(MLP(SBF(\alpha_{i},\alpha_{j}))\odot MLP(RBF(d_{ij})))
\end{equation}
where the formulas for RBF and SBF are 

\begin{equation}
\begin{aligned}
    RBF(d_{ij})&=j_l\left(\frac{\phi_{ln}}{C_e}d_{ij}\right),\\
 SBF(\alpha_{i}, \alpha_{j})&=Y_l^m(\alpha_{i}, \alpha_{j}).
\end{aligned}
\end{equation}
Here, $j_l$ is a spherical Bessel function of order $l$, and $\phi_{ln}$ is the $n$-th root of the $l$-order Bessel function. The edge neighbor cutoff is represented by $C_e$. $Y_l^m$ is a spherical harmonic function of degree $m$ and order $l$. 

Note that the existing methods use one embedding block to process the atom features and bond distances \cite{xie2018crystal,schutt2017schnet,chen2019graph,louis2020graph,yan2022periodic,lin2023efficient}. Our approach goes beyond by modeling atom features and structural information separately. There are two benefits of such design: (1) each embedding block can focus on one kind of information; (2) our approach can be easily extended to incorporate additional structural information.

\subsection{Message Passing Module}\label{MessagePassing}
Based on the outputs from the embedding module, the message passing module with several interaction blocks in series connection is designed. 
The message passing module aggregates the feature embedding and structure embedding of neighbors, which is shown as follows:
\begin{equation}\label{node_embedding}
    h_i^{(l)}=f^{(l)}(h_i^{(l-1)},h_j^{(l-1)},s_{i j})_{j\in NB_i},
\end{equation}
where $f^{(l)}$ represent the $l$-th interaction block, and $NB_i$ is the edge neighbors of atom $i$, which are determined by the edge neighbor cutoff value. The primary goal of the formula is to convolve the features of all neighboring atoms $j$ onto atom $i$.

\subsubsection{Interaction Block}\label{secmessage_passing}
The pipeline of interaction block is shown in Figure \ref{all_arc}(d), which shows the process of updating the feature embedding $h_i^{(l)}$ for atom $i$ is. 
The fusion of node embedding is computed as follows:
\begin{equation}
\begin{aligned}
    e_{ij}^{(l)}&=MLP((h_i^{(l)}\Vert h_j^{(l)}\Vert s_{ij})),\\
    e_{ij}^{(l)'}&=Active(e_i^{(l)}),\\
    t_i&=\sum_{j\in NB_i}(s_{ij}^{(l)'}\odot s_{ij}^{(l)}),\\
    h_i^{(l+1)}&=ReLu(h_i^{(l)}+LNorm(t_i)).
\end{aligned}    
\end{equation}
$e_{ij}^{(l)}$ represents the concatenated embedding of $h_i^{(l)}$, $h_j^{(l)}$, and $s_{ij}$, while $e_{ij}^{(l)'}$ denotes its activated form. Symbols $\odot$ and $\Vert$ respectively denote the Hadamard Product and concatenation operations. The variable $t_i$ aggregates all neighbor information of node $i$, serving as an intermediate aggregation. Ultimately, the updated node embedding $h_i^{(l+1)}$ is obtained by applying activation to the sum of $h_i^{(l)}$ and layer normalization to $t_i$.
Note that the first interaction block takes the atom embedding $h^{(0)}$ and structural embedding $s_{ij}$ as inputs.

\subsection{Readout Module}
Lastly, the readout module utilizes average pooling to compile and consolidate the atom features from all atoms within the unit cell. 
The final prediction is computed as
\begin{equation}\label{prediction}
    \hat{y}=MLP(\frac{1}{N}\sum_{n=1}^{N}h_n^{(L)}),
\end{equation}
where $h_n^{(L)}$ represents the final embedding acquired as atom n undergoes a total of $L$ interaction blocks. The overarching loss function is founded on the principle of minimizing the average error between predicted and actual values. This calculation method is widely adopted in related literature \cite{xie2018crystal,chen2019graph,louis2020graph,choudhary2021atomistic,lin2023efficient}.

Based on the $\hat{y}$, we introduce the loss of ADA-GNN.
We denote the parameters of ADA-GNN as $\theta$ for simplicity.
For each episode, $\theta$ is updated through gradient descent:
\begin{equation}\label{update}
\theta' = \theta - \lambda \nabla \mathbb{L}_\theta (f{(\theta)}),
\end{equation}
where $\lambda$ is the learning rate, and the $\mathbb{L}$ is the loss function of function $f(\theta)$ with respect to the parameter $\theta$.

The loss $\mathbb{L}_\theta$ is computed across the mini-batch:
\begin{equation}\label{loss}
\mathbb{L}_\theta(f{(\theta)}) = \frac{1}{B}\sum_{b=1}^B(\hat{y_b} - y_b)^2,
\end{equation}
where $B$ represents the batch size, $\hat{y_b}$ is the predicted value calculated from formula \eqref{prediction}, and $y_b$ is the ground truth.


\subsection{Training Algorithm and Time Cost Analysis}\label{secalgorithm}
\begin{algorithm}[t]
\caption{The Training Algorithm of ADA-GNN}
\label{alg1}
\LinesNumbered 
\KwIn{$\bm A = [a_1,..., a_N]^\top \in \mathbb{A}^N$, $\bm X = [\bm x_1, ..., \bm x_N]^\top \in \mathbb{R}^{N\times3}$, $\bm L = [\bm l_1, \bm l_2, \bm l_3]^\top \in \mathbb{R}^{3\times3}$} 
\KwOut{$\theta$} 

\medskip
\textbf{\emph{Part I: Graph Structure Acquisition}}

\For{$i=1$ to $N$}
{
    \For{$j=1$ \textbf{to} $m_i$}
    {
        Compute $d_{i j}$ via: $d_{i j} = ||\mathbf{x_j} - \mathbf{x_i}||_2$ \\
        \For{$k=1$ \textbf{to} $r_i$}
        {
            Get $\alpha_{i i_k}$ via: $\alpha_{i i_k} = angle(\overrightarrow{ii_k}, \vec{ij})$ \\
        }
        \For{$k=1$ \textbf{to} $r_j$}
        {
        Get $\alpha_{j j_k}$ via: $\alpha_{j j_k} = angle(\overrightarrow{jj_k}, \vec{ji})$ \\
        }
    }
}

\medskip
\textbf{\emph{Part II: Model Training }}

\While{not done}{
    \For{$b=1$ to $B$}{
        \For{$i=1$ to $N$}{
            Compute $h^0$ via formulate \eqref{equnode}.
            \For{$j=1$ \textbf{to} $m_i$}{
                Compute $emb_{ij}$ via formulate \eqref{edge_embedding}.
            }
        }
        \For{$i=1$ to $N$}{
            Update $h_i$ via formulate \eqref{node_embedding}\\
        }
        Compute $\hat{y}$ via formulate \eqref{prediction}
    }    
    Compute $\mathbb{L}_\theta(f_{(\theta)})$ via formulate \eqref{loss}
    Update $\theta$ via formulate \eqref{update}
}

\end{algorithm}
The training process of the ADA-GNN is outlined in Algorithm \ref{alg1}. As the ADA-GNN is specially designed based on the $G_M$, we incorporate the calculation of $G_M$ as part of the training process. Thus, the training algorithm has two phases: the calculation of $G_M$ and the training of ADA-GNN.


The first phase involves computing node attributes, edge attributes, and angle attributes. 
The distance $d$ between atoms is directly calculated using the formula in the third row.
Following this, lines 4-9 of the algorithm illustrate the calculation of the angle $\alpha$.
Once all the inputs are gathered, we have the second phase. Within our architecture, rows 16-18 handle the computation of structural attributes, while row 15 deals with node features. The message passing operation within rows 20-22 amalgamates structural attributes and neighboring information, consequently updating node attributes. In row 23, predictions are generated utilizing the updated node attributes. Subsequently, the loss function is computed in lines 25 and 26, respectively, and the network parameters are updated accordingly. This entire training process iterates until reaching the maximum number of training rounds.

After training, the inference time for crystal materials is strongly affected by the volume of the inputs. Thus, we provide an analysis of the input volume of ADA-GNN. 
The volume of bond distance is $O(NM)$, where N is the number of atoms in the crystal cell, and M is the average number of edge neighbors for each atom. On the other hand, the volume of angle information is $O(NMK)$, where $K$ is the average number of angle neighbors for each node. Notably, as per Formula \eqref{equneigh}, the number of edge length neighbors $M$ is a square multiple of the number of angle neighbors $K$. Consequently, the overall volume of both edge and angle information is  $O(NM\log M)$.
As the ALiGNN is the only one considering the bond angle information, we take ALiGNN as a comparison. The volume of inputs in ALiGNN is $O(NM^2)$. Thus, we make a notable advancement, as compared to the $O(NM^2)$ with one kind of cutoff, signifying a reduction in computational complexity by one logarithmic dimension. The inference time is also compared in Section 5.3.

\section{Experiments}

The following research questions guide the remainder of the paper: 
(\textbf{Q1}) can our proposed ADA-GNN outperform SOTA baselines; 
(\textbf{Q2}) is there any improvement in prediction time compared to similar algorithms; 
(\textbf{Q3}) how does angel information affect the performance?
\subsection{Experimental Setup}


The evaluation of ADA-GNN's efficacy and inference speed spans across two extensive material benchmark datasets: The Materials Project-2018.6 \cite{jain2013materials} and JARVIS \cite{choudhary2020joint}.
The number of data entries corresponding to each dataset is presented in Table \ref{tab:Dataset}, enhancing the transparency and consistency of our evaluations against established benchmarks.
Seven representative baseline methods are selected: CGCNN, SchNet, MEGNET, GATGNN, ALiGNN, Matformer, and PotNet. 
The main differences between each algorithm are shown in the Table \ref{tab:difference}. 

Our computational setup involves the utilization of a single NVIDIA GeForce RTX 24G 3090 GPU for all tasks. In terms of implementation, all models undergo training using the Adam optimizer \cite{kingma2014adam} with a one-cycle learning rate scheduler \cite{smith2019super}, and the learning rate is configured at 0.001, with a batch size of 64 and a training epoch count of 500 serving as our chosen training parameters.

\begin{table}[t]
  \centering  
  \begin{threeparttable}  
  \caption{Statistics of datasets.}  
  \label{tab:Dataset} 
    \begin{tabular}{c|c|ccc}
    \toprule  
    Dataset&Tasks&\# train&\# valid&\# test\cr
    \midrule  
    \multirow{4}{*}{JARVIS}
    &F. Energy&44578&5572&5572\cr
    &T. energy&44578&5572&5572\cr
    &B. Gap MBJ&14537&1817&1817 \cr
    &Ehull&44296&5537&5537\cr
    \midrule
    Materials&
    F. Energy&60000&5000&4239\cr
    Project&
    B. Gap&60000&5000&4239\cr 
    \bottomrule  
    \end{tabular}  
    \end{threeparttable}  
\end{table}

\subsection{Experimental Results}
\begin{table*}[h]  
  \centering  
  \begin{threeparttable}  
  \caption{Comparison between our method and other baselines in terms of test MAE on JARVIS dataset and The Materials Project dataset. The best results are shown in {\bf bold}.
  The last line shows the improvement rate compared to PotNet.}
  \label{tab:Jarvis and MP} 
    \begin{tabular}{c|cccccc}
    \toprule  
    \multirow{3}{*}{Method}&
    \multicolumn{4}{c|}{JARVIS}&\multicolumn{2}{c}{Materials Project}\cr
    \cmidrule(lr){2-7}
    &\multicolumn{1}{c}{Formation Energy}&\multicolumn{1}{c}{Total energy}&\multicolumn{1}{c}{Bandgap(MBJ)}&\multicolumn{1}{c}{Ehull}&\multicolumn{1}{c}{Formation Energy}&\multicolumn{1}{c}{Band Gap}\cr  
    \cmidrule(lr){2-7}     
    &eV/atom&eV/atom&eV&eV&eV/atom&eV\cr  
    \midrule  
    CGCNN&0.063&0.078&0.41&0.17&0.031&0.292\cr
    SchNet&0.045&0.047&0.43&0.14&0.033&0.345\cr
    MEGNET&0.047&0.058&0.34&0.084&0.030&0.307\cr
    GATGNN&0.047&0.056&0.51&0.12&0.033&0.280\cr
    ALiGNN&0.0331&0.037&0.31&0.076&0.0221&0.218\cr
    Matformer&0.0325&0.035&0.30&0.064&0.0210&0.211\cr
    PotNet& {0.0294}&{0.032}&{0.27}& {0.055}&{0.0188}&{0.204}\cr
    \midrule
    \multirow{2}{*}{ADA-GNN}&\bf 0.0288 &\bf 0.031&\bf 0.25&\bf 0.043&\bf 0.0182&\bf 0.197\cr  
    &($ 2.04\%\uparrow$)&($3.13\%\uparrow$)&($7.41\%\uparrow$)&($21.82\%\uparrow$)&($ 3.19\%\uparrow$)&($3.43\%\uparrow$)\cr
    \bottomrule  
    \end{tabular}  
    \end{threeparttable}  
\end{table*}
In this section, our primary goal is to respond to Q1 by conducting a comprehensive assessment of the ADA-GNN's performance.
The evaluation metric employed here is the test Mean Absolute Error (MAE), consistent with prior studies \cite{choudhary2021atomistic,lin2023efficient,yan2022periodic,xie2018crystal}. 
As outlined in Table \ref{tab:Jarvis and MP}, ADA-GNN showcases remarkable performance across all tasks on both benchmark datasets. 
When benchmarked against the latest state-of-the-art method, PotNet \cite{lin2023efficient}, our algorithm achieves a notable performance improvement of $2.04\%$ to $21.82\%$. This underscores the effectiveness of our network architecture in fully harnessing the role of bond angles, achieving performance levels unattainable solely through the utilization of bond distances.

Specifically, our approach demonstrates superior performance, with enhancements ranging from $9.63\%$ to $43.32\%$  across diverse tasks compared with ALiGNN \cite{choudhary2021atomistic}. 
While both our architecture ADA-GNN and the ALiGNN approach incorporate angle information, there are two notable differences between our ADA-GNN and ALiGNN. Firstly, our network employs a dual-scale cutoff, effectively leveraging less critical angle information compared to ALiGNN. Secondly, there is another difference that lies in the construction of the structure embedding: ALiGNN uses one embedding block to incorporate the information of edge length, edge angle, and node features. In contrast, we introduce the node embedding block and structure embedding block that segregates structural embeddings from node embeddings, thereby mitigating the impact of node feature updates on structural attributes.
The results reveal that our network utilizes less angle information than the ALiGNN while outperforming it. This outcome underscores the robust effectiveness of our proposed ADA-GNN architecture.

\subsection{Prediction Time Cost Analysis}\label{sectime}
\begin{table}[ht]  
  \centering  
  \begin{threeparttable}  
  \caption{Prediction time compared with Matformer, ALiGNN, and PotNet on JARVIS formation energy prediction.
  } 
  \label{tab:Training time} 
    \begin{tabular}{ccc}
    \toprule  
    \multirow{2}{*}{Method}
    &\multicolumn{2}{c}{Prediction Time}\cr
    \cmidrule(lr){2-3}
    &Total Prediction Time&Per Crystal Time\cr   
    \midrule  
    Matformer&2.05min&23ms\cr
    Potnet&1.08min&12ms\cr  
    \midrule
    \midrule
    ALiGNN&5.41min&61ms\cr
    ADA-GNN& 1.8min&19ms\cr  
    \bottomrule  
    \end{tabular}  
    \end{threeparttable}  
\end{table}

In this section, we aim to answer question Q2 by analyzing the prediction time cost of ADA-GNN. 
In the realm of new material discovery, employing machine learning algorithms for crystal property prediction primarily aims to expedite the screening process for newly generated crystal stabilities \cite{merchant2023scaling}. Consequently, beyond the precision of crystal property predictions, the time invested in forecasting these properties serves as a crucial metric for assessing the efficacy of the methodology.
 
Based on this, we compare the predicating time using the JARVIS formation energy dataset with the most related method, ALiGNN, which also computes angles, and the most recent method, MatFormer, and PotNet. 
The column labeled "Total Prediction Time" denotes the cumulative time expended in predicting 5572 molecules within the test set, while, the "Per Crystal Time" column articulates the average time allocated for predicting each individual crystal molecule.

Compared with ALiGNN, our ADA-GNN is more efficient. As shown in Table \ref{tab:Training time}, our ADA-GNN is nearly three times faster in inference time for the whole test set. This result indicates that our dual scale neighbor acquisition mechanism can significantly reduce prediction time, thereby meeting the needs of downstream tasks.
Compared to methods exclusively targeting bond distance, our algorithm sustains competitive prediction times. While ADA-GNN falls behind the efficiency-focused bond distance-based PotNet, it outperforms PotNet in performance. Notably, our method excels in prediction time compared to MatFormer, another bond distance-focused approach, underscoring the efficiency gains achieved by our algorithm.
\subsection{The Affection of Missing Angle Information}
\begin{table}[t]  
  \centering  
  \begin{threeparttable}  
  \caption{Ablation studies for the effects of ignoring angles.} 
  \label{tab:Ablation studies}  
    \begin{tabular}{c|cc}
    \toprule  
    Tasks&Ours&Ours without angles\cr
    \midrule  
    F. Energy (eV/atom)&\bf 0.0288&0.0301 ($4.51\%\downarrow$)\cr
    T. energy (eV/atom)&\bf 0.031&0.033 ($6.45\%\downarrow$)\cr
    B. Gap(MBJ) (eV)&\bf 0.25&0.32 ($28.0\%\downarrow$)\cr
    Ehull (eV)&\bf 0.043&0.070 ($62.8\%\downarrow$)\cr
    F. Energy (eV/atom)&\bf 0.0182&0.0189 ($3.85\%\downarrow$)\cr
    B. Gap (eV)&\bf 0.197&0.200 ($1.52\%\downarrow$)\cr 
    \bottomrule  
    \end{tabular}  
    \end{threeparttable}  
\end{table}
In this section, our objective is to address Q3 by emphasizing the pivotal role played by angle information within the ADA-GNN. We conduct experiments on The Materials Project-2018.6 and JARVIS tasks, employing test Mean Absolute Error (MAE) as the evaluation metric.

To underscore the significance of angle information, we compare the ADA-GNN model with `Ours without angles', a variant that exclusively considers distances while disregarding angle information. 
The initial structure embedding in `Ours without angles' is computed as the following formula :
\begin{equation} \label{withoutangle}
    s_{ij}=MLP(MLP(RBF(d_{ij}))).
\end{equation}
For a fair comparison, the two architectures have the same settings for all tasks except for structural embedding.

As illustrated in Table \ref{tab:Ablation studies}, excluding angle information results in a performance decline, reducing the accuracy by a range of $1.52\%$ to $62.8\%$. This stark contrast underscores the pivotal role that angle information plays within the representation, highlighting its impact on predictive accuracy.

\section{Conclusion}

The infinite bond angle problem hinders the utilization of bond angle in modeling crystal materials. To alleviate this issue, we propose a dual scale modeling in selecting neighbors:  a larger scale truncation for edge neighbors and a smaller scale truncation for angle neighbors. 
Then, we propose a novel approach named ADA-GNN for property prediction tasks based on the dual scale modeling. The atom features and structure information are processed independently, which can enhance training stability and improve the accuracy of model predictions. 
Extensive experiments are conducted on two large-scale real-world datasets compared with the seven graph-learning-based methods.
The results show that the ADA-GNN achieves state-of-the-art (SOTA) results.
Furthermore, the ADA-GNN is 3.21 times faster than ALiGNN in terms of inference time per crystal.

\bibliographystyle{unsrt}  
\bibliography{references}

\end{document}